\documentclass[10pt,final]{article}

\usepackage{eacl2009}
\usepackage{times}
\usepackage{latexsym}
\usepackage[fleqn]{amsmath}
\usepackage{amssymb}

\usepackage[utf8x]{inputenc}
\usepackage{graphicx}
\usepackage{color}
\usepackage{tabularx}
\usepackage{t1enc}
\usepackage[T1]{url}

\usepackage{abbrevs}
\let\em\itswitch

\newabbrev\fn{FrameNet}
\newabbrev\tb{TreeBank}
\newabbrev\pb{PropBank}
\newabbrev\nb{NomBank}
\newabbrev\nlg{Natural Language Generation}[NLG]

\newif\ifpdf
\ifx\pdfoutput\undefined \else
  \ifx\pdfoutput\relax
  \else
    \ifcase\pdfoutput
    \else
      \pdftrue
    \fi
  \fi
\fi

\providecommand{\ie}{\textit{i.e.} }%
\providecommand{\eg}{\textit{e.g.} }%
\catcode`\=13
\def{$\bowtie$}

\title{What's in a Message?}

\author{Stergos D. Afantenos \and Nicolas Hernandez \\
LINA, (UMR CNRS 6241) \\
Universit\'e de Nantes, France \\
\small\url{stergos.afantenos@univ-nantes.fr} \\
\small\url{nicolas.hernandez@univ-nantes.fr}}

\date{}

\begin{document}

\maketitle

\begin{abstract}
In this paper we present the first step in a larger series of experiments for the induction of predicate/argument structures. The structures that we are inducing are very similar to the conceptual structures that are used in Frame Semantics (such as \fn). Those structures are called messages and they were previously used in the context of a multi-document summarization system of evolving events. The series of experiments that we are proposing are essentially composed from two stages. In the first stage we are trying to extract a representative vocabulary of words. This vocabulary is later used in the second stage, during which we apply to it various clustering approaches in order to identify the clusters of predicates and arguments---or frames and semantic roles, to use the jargon of Frame Semantics. This paper presents in detail and evaluates the first stage.
\end{abstract}

\section{Introduction}
Take a sentence, any sentence for that matter; step back for a while and try to perceive that sentence in its most abstract form. What you will notice is that once you try to abstract away sentences, several regularities between them will start to emerge. To start with, there is almost always an \emph{action} that is performed.\footnote{In linguistic terms, an action-denoting word is also known as a \emph{predicate}.} Then, there is most of the times an \emph{agent} that is performing this action and a \emph{patient} or a \emph{benefactor} that is receiving this action, and it could be the case that this action is performed with the aid of a certain \emph{instrument}. In other words, within a sentence---and in respect to its action-denoting word, or predicate in linguistic terms---there will be several entities that are associated with the predicate, playing each time a specific \emph{semantic role}.

The notion of semantic roles can be traced back to Fillmore's \shortcite{Fillmore.76} theory of Frame Semantics. According to this theory then, a \emph{frame} is a conceptual structure which tries to describe a stereotypical situation, event or object along with its participants and props. Each frame takes a name (\eg \textsc{Commercial Transaction}) and contains a list of \emph{Lexical Units (LUs)} which actually evoke this frame. An LU is nothing else than a specific word or a specific meaning of a word in the case of polysemous words. To continue the previous example, some LUs that evoke the frame of \textsc{Commercial Transaction} could be the verbs \texttt{buy}, \texttt{sell}, etc. Finally, the frames contain several frame elements or \emph{semantic roles} which actually denote the abstract conceptual entities that are involved with the particular frame.

Research in semantic roles can be distinguished into two major branches. The first branch of research consists in \emph{defining} an ontology of semantic roles, the frames in which the semantic roles are found as well as defining the LUs that evoke those frames. The second branch of research, on the other hand, \emph{stipulates} the existence of a set of frames, including semantic roles and LUs; its goal then, is the creation of an algorithm that given such a set of frames containing the semantic roles, will be able to label the appropriate portions of a sentence with the corresponding semantic roles. This second branch of research is known as \emph{semantic role labeling}.

Most of the research concerning the definition of the semantic roles has been carried out by linguists who are \emph{manually} examining a certain amount of frames before finally defining the semantic roles and the frames that contain those semantic roles. Two such projects that are widely known are the \fn \cite{Baker&al.98,FramenetBook} and \pb/\nb\footnote{We would like to note here that although the two approaches (\fn and \pb/\nb) share many common elements, they have several differences as well. Two major differences, for example, are the fact that the Linguistic Units (\fn) are referred to as Relations (\pb/\nb), and that for the definition of the semantic roles in the case of \pb/\nb there is no reference ontology. A detailed analysis of the differences between \fn and \pb/\nb would be out of the scope of this paper.} \cite{Palmer&al.05,Meyers&al.04}. Due to the fact that the aforementioned projects are accompanied by a large amount of annotated data, computer scientists have started creating algorithms, mostly based on statistics \cite{Gildea&Jurafsky.02,Xue.08} in order to automatically label the semantic roles in a sentence. Those algorithms take as input the frame that contains the roles as well as the predicate\footnote{In the case of \fn the predicate corresponds to a ``Linguistic Unit'', while in the case of \pb/\nb it corresponds to what is named ``Relation''.} of the sentence.

Despite the fact that during the last years we have seen an increasing interest concerning semantic role labeling,\footnote{\textit{Cf}, for example, the August 2008 issue of the journal \textit{Computational Linguistics} (34:2).} we have not seen many advancements concerning the issue of \emph{automatically inducing} semantic roles from raw textual corpora. Such a process of induction would involve, firstly the identification of the words that would serve as predicates and secondly the creation of the appropriate clusters of word sequences, within the limits of a sentence, that behave similarly in relation to the given predicates. Although those clusters of word sequences could not actually be said to serve in themselves as the semantic roles,\footnote{At least as the notion of semantic roles is proposed and used by \fn.} they can nevertheless be viewed as containing characteristic word sequences of specific semantic roles. The last point has the implication that if one is looking for a human intuitive naming of the semantic role that is implied by the cluster then one should look elsewhere. This is actually reminiscent of the approach that is carried out by \pb/\nb in which each semantic role is labeled as $Arg1$ through $Arg5$ with the semantics given aside in a human readable natural language sentence.

Our goal in this paper is to contribute to the research problem of frame induction, that is of the creation of frames, including their associated semantic roles, given as input only a set of textual documents. More specifically we propose a general methodology to accomplish this task, and we test its first stage which includes the use of corpus statistics for the creation of a subset of words, from the initial universe of initial words that are present in the corpus. This subset will later be used for the identification of the predicates as well as the semantic roles. Knowing that the problem of frame induction is very difficult in the general case, we limit ourselves to a specific genre and domain trying to exploit the characteristics that exist in that domain. The domain that we have chosen is that of the terroristic incidents which involve hostages. Nevertheless, the same methodology could be applied to other domains.

The rest of the paper is structured as follows. In section~\ref{sec:data} we describe the data on which we have applied our methodology, which itself is described in detail in section~\ref{sec:method}.  Section~\ref{sec:experiments_results} describes the actual experiments that we have performed and the results obtained, while a discussion of those results follows in section~\ref{sec:discussion}. Finally, section~\ref{sec:related_work} contains a description of the related work while we present our future work and conclusions in section~\ref{sec:conclusions}.

\section{The Annotated Data}\label{sec:data}
The annotated data that we have used in order to perform our experiments come from a previous work on automatic multi-document summarization of events that evolve through time \cite{Afantenos&al.08:JIIS,Afantenos&al.05:NLUCS,Afantenos&al.04:SETN}.
The methodology that is followed is based on the identification of similarities and differences---between documents that describe the evolution of an event---synchronically as well as diachronically. In order to do so, the notion of \emph{Synchronic and Diachronic} cross document Relations (SDRs),\footnote{Although a full analysis of the notion of Synchronic and Diachronic Relations is out of the scope of this paper, we would like simply to mention that the premises on which those relations are defined are similar to the ones which govern the notion of \emph{Rhetorical Structure Relations} in Rhetorical Structure Theory (RST) \cite{Taboada&Mann:RST1}, with the difference that in the case of SDRs the relations hold across documents, while in the case of RSTs the relation hold inside a document.} was introduced. SDRs connect not the documents themselves but some semantic structures that were called \emph{messages}. The connection of the messages with the SDRs resulted in the creation of a semantic graph that was then fed to a \nlg (NLG) system in order to produce the final summary. Although the notion of messages was originally inspired by the notion of messages as used in the area of \nlg, for example during the stage of \emph{Content Determination} as described in \cite{Reiter&Dale97}, and in general they do follow the spirit of the initial definition by Reiter \& Dale, in the following section we would like to make it clear what the notion of messages represents for us. In the rest of the paper, when we refer to the notion of messages, it will be in the context of the discussion that follows.

\subsection{Messages}\label{sec:data:msgs}
The intuition behind messages, is the fact that during the evolution of an event we have several activities that take place and each activity is further decomposed into a series of \emph{actions}. Messages were created in order to capture this abstract notion of actions. Of course, actions usually implicate several entities. In this case, entities were represented with the aid of a domain ontology. Thus, in more formal terms a message $m$ can be defined as follows:
\begin{center}\small
\begin{tabular}{c}
  $m$\texttt{ = message\_type (arg$_1$, $\ldots$ , arg$_n$)}\\
  where \texttt{arg}$_i$ $\in$ Topic Ontology, $i \in \{1,\ldots,n\}$
\end{tabular}
\end{center}
In order to give a simple example, let us take for instance the case of the hijacking of an airplane by terrorists. In such a case, we are interested in knowing if the airplane has arrived to its destination, or even to another place. This action can be captured by a message of type \verb|arrive| whose arguments can be the entity that arrives (the airplane in our case, or a vehicle, in general) and the location that it arrives. The specifications of such a message can be expressed as follows:
\begin{center}
\ttfamily\small
arrive (what, place)\\
\begin{tabularx}{1.6in}{rX}
\small  what  :& Vehicle\\
\small  place :& Location\\
\end{tabularx}
\end{center}
The concepts \texttt{Vehicle} and \texttt{Location} belong to the ontology of the topic; the concept \texttt{Air\-pla\-ne} is a sub-concept of the \texttt{Vehicle}. A sentence that might instantiate this message is the following:
\begin{quote}
  The Boeing 747 arrived at the airport of Stanstend.
\end{quote}
The above sentence instantiates the following me\-ssa\-ge:
\begin{quote}
\small\texttt{arrive ("Boeing 747", "airport of Stanstend")}
\end{quote}

The domain which was chosen was that of terroristic incidents that involve hostages. An empirical study, by three people, of 163 journalistic articles---written in Greek---that fell in the above category, resulted in the definition of 48 different message types that represent the most important information in the domain. At this point we would like to stress that what we mean by ``most important information'' is the information that one would normally expect to see in a typical summary of such kinds of events. Some of the messages that have been created are shown in Table~\ref{table:msgs}; figure~\ref{fig:msgSpecs} provides full specifications for two messages.

\begin{table}[htb]
  \ttfamily\small\centering
  \begin{tabular}{|l|l|}
    \hline
    free & explode\\
    \hline
    kill & kidnap \\
    \hline
    enter & arrest\\
    \hline
    negotiate & encircle\\
    \hline
    escape\_from & block\_the\_way\\
    \hline
    give\_deadline & \\
    \hline
  \end{tabular}
  \rmfamily\normalsize
  \caption{Some of the message types defined.}\label{table:msgs}
\end{table}

\begin{figure}[htb]
\centering
\begin{tabular}{|l|}
\hline
\textbf{negotiate} (who, with\_whom, about)\\
  \quad who : Person\\
  \quad with\_whom : Person\\
  \quad about : Activity\\
\hline
\textbf{free} (who, whom, from)\\
  \quad who : Person\\
  \quad whom : Person\\
  \quad from : Place $\vee$ Vehicle\\
\hline
\end{tabular}
  \caption{An example of message specifications}
  \label{fig:msgSpecs}
\end{figure}

Although in an abstract way the notion of messages, as presented in this paper approaches the notion of frame semantics---after all, both messages and frame semantics are concerned with ``who did what, to whom, when, where and how''---it is our hope that our approach could ultimately be used for the problem of frame induction. Nevertheless, the two structures have several points in which they differ. In the following section we would like to clarify those points in which the two differ.

\subsection{How Messages differ from Frame Semantics}\label{sec:msg_diff_sr}
As it might have been evident until now, the notions of messages and frame semantics are quite similar, at least from an abstract point of view. In practical terms though, the two notions exhibit several differences.

To start with, the notion of messages has been used until now only in the context of automatic text summarization of multiple documents. Thus, the aim of messages is to capture the \emph{essential information} that one would expect to see in a typical summary of this domain.\footnote{In this sense then, the notion of messages is reminiscent of Schank \& Abelson's \shortcite{Schank&Abelson.77} notion of \emph{scripts}, with the difference that messages are not meant to exist inside a structure similar to Schank \& Abelson's ``scenario''. We would like also to note that the notion of messages shares certain similarities with the notion of \emph{templates} of Information Extraction, as those structures are used in conferences such as MUC. Incidentally, it is not by chance that the ``M'' in MUC stands for Message (Understanding Conference).} In contrast, semantic roles and the frames in which they exist do not have this limitation.

Another differentiating characteristic of frame semantics and messages is the fact that semantic roles always get instantiated within the boundaries of the sentence in which the predicate exists. By contrast, in messages although in the vast majority of the cases there is a one-to-one mapping from sentences to messages, in some of the cases the arguments of a message, which correspond to the semantic roles, are found in neighboring sentences. The overwhelming majority of those cases (which in any case were but a few) concern \emph{referring expressions}. Due to the nature of the machine learning experiments that were performed, the actual entities were annotated as arguments of the messages, instead of the referring expressions that might exist in the sentence in which the message's predicate resided.

A final difference that exists between messages and frame semantics is the fact that messages were meant to exist within a certain domain, while the definition of semantic roles is usually independent of a domain.\footnote{We would like to note at this point that this does not exclude of course the fact that the notion of messages could be used in a more general, domain independent way. Nevertheless, the notion of messages has for the moment been applied in two specific domains \cite{Afantenos&al.08:JIIS}.}

\section{The Approach Followed}\label{sec:method}
\begin{figure*}[htb]
\begin{center}
  \includegraphics[width=\textwidth]{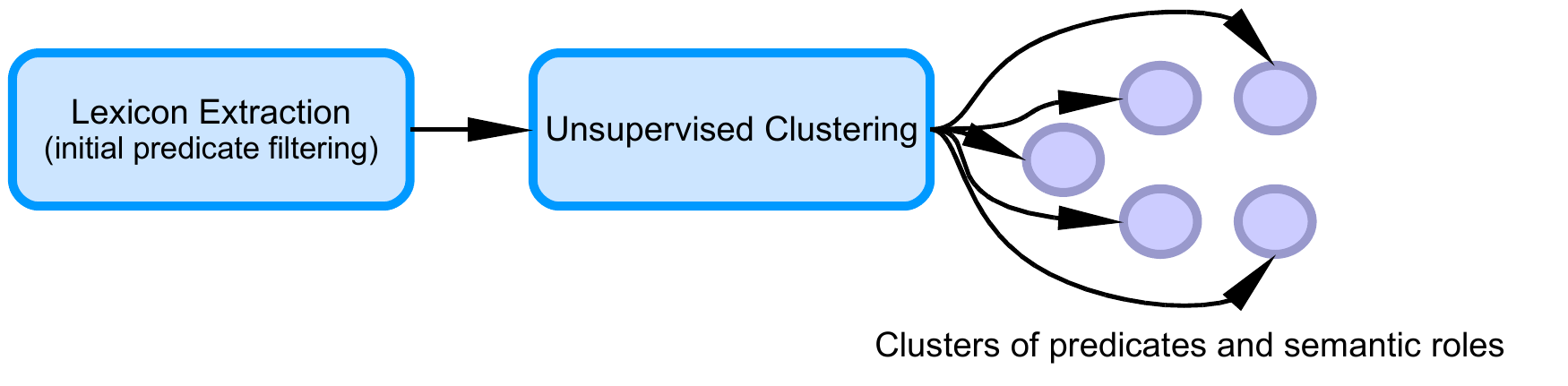}
  \caption{Two different stages in the process of predicate clustering} \label{fig:stages}
\end{center}
\end{figure*}

A schematic representation of our approach is shown in Figure~\ref{fig:stages}. As it can be seen from this figure, our approach comprises two stages. The first stage concerns the creation of a lexicon which will contain as most as possible---and, of course, as accurately as possible---candidates that are characteristic either of the predicates (message types) or of the semantic roles (arguments of the messages). This stage can be thought of as a filtering stage. The second stage involves the use of unsupervised clustering techniques in order to create the final clusters of words that are characteristic either of the predicates or of the semantic roles that are associated with those predicates. The focus of this paper is on the first stage.

As we have said, our aim in this paper is the use of statistical measures in order to extract from a given corpus a set of words that are most \emph{characteristic} of the messages that exist in this corpus. In the context of this paper, a word will be considered as being characteristic of a message if this word is employed in a sentence that has been annotated with that message. If a particular word does not appear in any message annotated sentence, then this word will not be considered as being characteristic of this message. In more formal terms then, we can define our task as follows. If by $\mathcal{U}$ we designate the set of all the words that exist in our corpus, then we are looking for a set $\mathcal{M}$ such that:
\begin{multline}\label{eq}
\mathcal{M} \subset \mathcal{U} \quad \wedge  \\
\quad w \in \mathcal{M} \Leftrightarrow m \textrm{ appears at least once} \\
\textrm{ in a message instance}
\end{multline}
In order to extract the set $\mathcal{M}$ we have employed the following four statistical measures:
\begin{description}
  \item[Collection Frequency:] The set that results from the union of the $n\%$ most frequent words that appear in the corpus.
  \item[Document Frequency:] The set that results from the union of the $n\%$ most frequent words of each document in the corpus.
  \item[tf.idf:] For each word in the corpus we calculate its $tf.idf$. Then we create a set which is the union of words with the highest $n\%$ \ $tf.idf$ score in each document.
  \item[Inter-document Frequency:] A word has inter-do\-cu\-ment frequency $n$ if it appears in at least $n$ documents in the corpus. The set with inter-document frequency $n$ is the set that results from the union of all the words that have inter-document frequency~$n$.
\end{description}
As we have previously said in this paper, our goal is the exploitation of the characteristic vocabulary that exists in a specific genre and domain in order to ultimately achieve our goal of message induction, something which justifies the use of the above statistical measures.
The first three measures are known to be used in context of Information retrieval to capture topical informations. The latter measure has been proposed by \cite{hernandezAl03} in order to extract rhetorical indicator phrases from a genre dependant corpus.

In order to calculate the aforementioned statistics, and create the appropriate set of words, we ignored all the stop-words. In addition we worked only with the \emph{verbs} and \emph{nouns}. The intuition behind this decision lies in the fact that the created set will later be used for the identification of the predicates and the induction of the semantic roles. As Gildea \& Jurafsky \shortcite{Gildea&Jurafsky.02}---among others---have mentioned, predicates, or action denoting words, are mostly represented by verbs or nouns.\footnote{In some rare cases predicates can be represented by adjectives as well.} Thus, in this series of experiments we are mostly focusing in the extraction of a set of words that approaches the set that is obtained by the union of all the verbs and nouns found in the annotated sentences.

\section{Experiments and Results}\label{sec:experiments_results}
The corpus that we have consists of 163 journalistic articles which describe the evolution of five different terroristic incidents that involved hostages. The corpus was initially used in the context of training a multi-document summarization system. Out of the 3,027 sentences that the corpus contains, about one third (1,017 sentences) were annotated with the 48 message types that were mentioned in section~\ref{sec:data:msgs}.

\begin{table}[htb]
  \small\centering
  \begin{tabular}{|l|l|}
    \hline
    \textbf{Number of Documents:} & 163\\
    \hline
    \textbf{Number of Token:} & 71,888\\
    \hline
    \textbf{Number of Sentences:} & 3,027\\
    \hline
    \textbf{Annotated Sentences (messages):} & 1,017\\
    \hline
    \textbf{Distinct Verbs and Nouns in the Corpus:} & 7,185\\
    \hline
    \textbf{Distinct Verbs and Nouns in the Messages:} & 2,426\\
    \hline
  \end{tabular}
  \rmfamily\normalsize
  \caption{Corpus Statistics.}\label{table:stats}
\end{table}

The corpus contained 7,185 distinct verbs and nouns, which actually constitute the $\mathcal{U}$ of the formula~(\ref{eq}) above. Out of those 7,185 distinct verbs and nouns 2,426 appear in the sentences that have been annotated with the messages. Our goal was to create this set that approached as much as possible to the set of 2,426 distinct verbs and nouns that are found in the messages.

Using the four different statistical measures presented in the previous section, we tried to reconstruct that set. In order to understand how the statistical measures behaved, we varied for each one of them the value of the threshold used. For each statistical measure used, the threshold represents something different. For the Collection Frequency measure the threshold represents the $n\%$ most frequent words that appear in the corpus. For the Document Frequency it represents the $n\%$ most frequent words that appear in each document separately. For \textit{tf.idf} it represents the words with the highest $n\%$ \textit{tf.idf} score in each document. Finally for the Inter-document Frequency the threshold represents the verbs and nouns that appear in at least $n$ documents. Since for the first three measures the threshold represents a percentage, we varied it from 1 to 100 in order to study how this measure behaves. For the case of the Inter-document Frequency, we varied the threshold from 1 to 73 which represents the maximum number of documents in which a word appeared.

In order to measure the performance of the statistical measures employed, we used four different evaluation measures, often employed in the information retrieval field. Those measures are the \emph{Precision, Recall, F-measure} and \emph{Fallout}. Precision represents the percentage of the correctly obtained verbs and nouns over the total number of obtained verbs and nouns. Recall represents the percentage of the obtained verbs and nouns over the target set of verbs and nouns. The F-measure is the harmonic mean of Precision and Recall. Finally, fallout represents the number of verbs and nouns that were wrongly classified by the statistical measures as belonging to a message, over the total number of verbs and nouns that do not belong to a message. In an ideal situation one expects a very high precision and recall (and  by consequence F-measure) and a very low Fallout.

\begin{figure*}[htb]
\begin{center}
    \includegraphics[width=\textwidth]{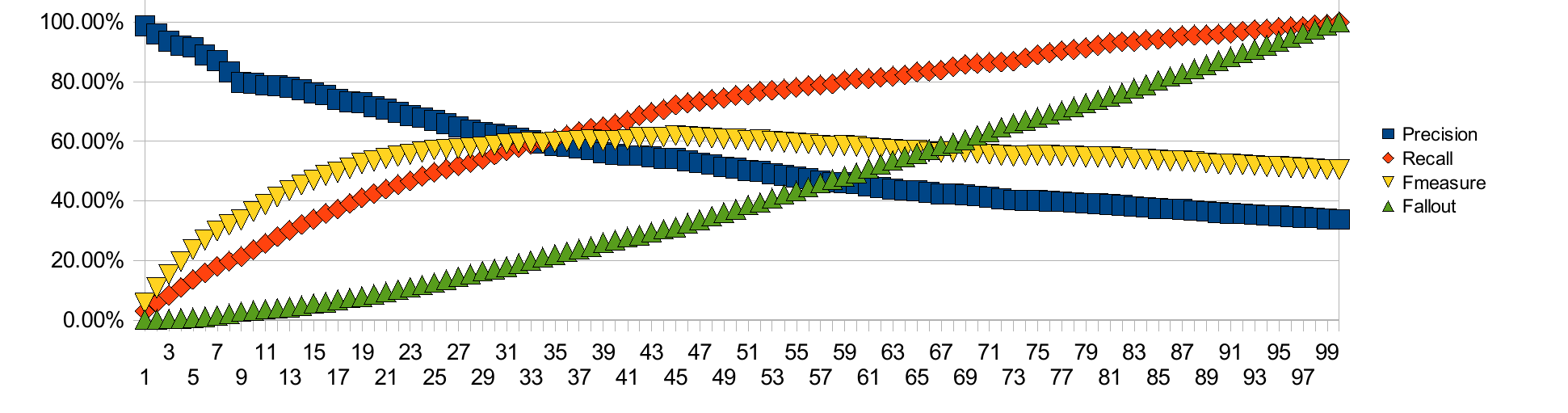}
  \caption{Collection Frequency statistics}\label{fig:cf}
\end{center}
\end{figure*}

\begin{figure*}[htb]
\begin{center}
    \includegraphics[width=\textwidth]{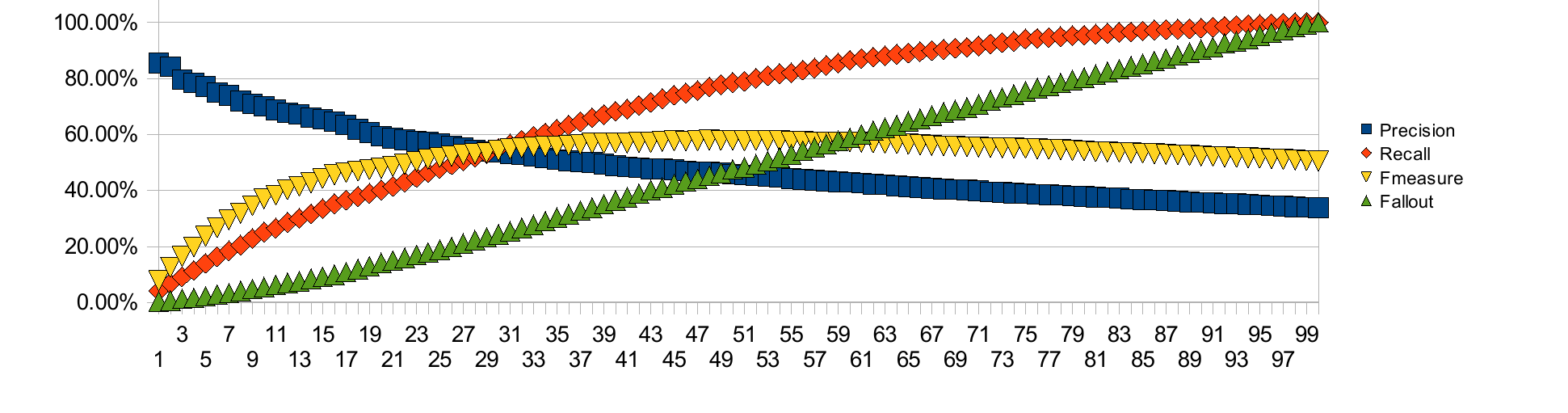}
  \caption{Document Frequency statistics}\label{fig:df}
\end{center}
\end{figure*}

\begin{figure*}[htb]
\begin{center}
    \includegraphics[width=\textwidth]{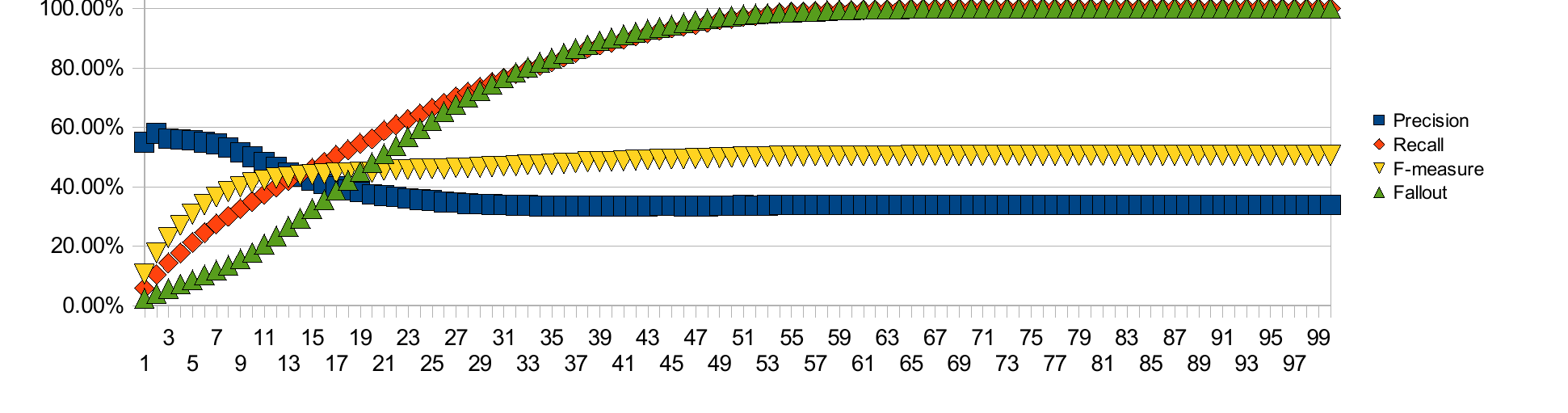}
  \caption{Tf.idf statistics}\label{fig:tfidf}
\end{center}
\end{figure*}

\begin{figure*}[htb]
\begin{center}
    \includegraphics[width=\textwidth]{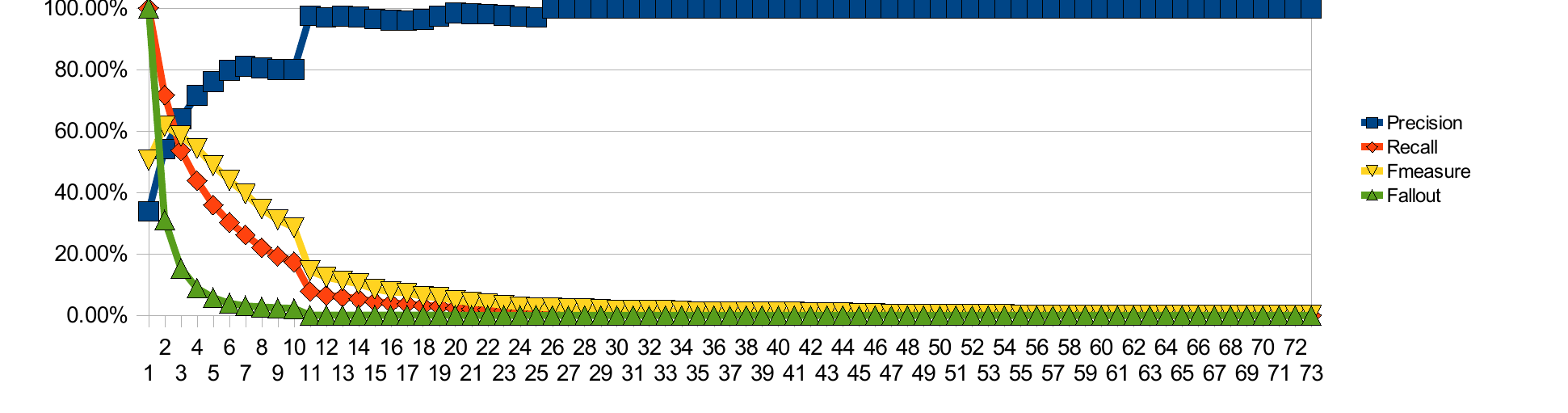}
  \caption{Inter-document frequency statistics}\label{fig:fit}
\end{center}
\end{figure*}

The obtained graphs that combine the evaluation results for the four statistical measures presented in section~\ref{sec:method} are shown in Figures~\ref{fig:cf} through~\ref{fig:fit}. A first remark that we can make in respect to those graphs is that concerning the collection frequency, document frequency and tf.idf measures, for small threshold numbers we have more or less high precision values while the recall and fallout values are low. This implies that for smaller threshold values the obtained sets are rather small, in relation to $\mathcal{M}$ (and by consequence to $\mathcal{U}$ as well). As the threshold increases we have the opposite situation, that is the precision falls while the recall and the fallout increases, implying that we get much bigger sets of verbs and nouns.

In terms of absolute numbers now, the best F-measure is given by the Collection Frequency measure with a threshold value of 46\%. In other words, the best results---in terms of F-measure---is given by the union of the 46\% most frequent verbs and nouns that appear in the corpus. For this threshold the Precision is 54.14\%, the Recall is 72.18\% and the F-measure is 61,87\%. This high F-measure though comes at a certain cost since the Fallout is at 31.16\%. This implies that although we get a rather satisfying score in terms of precision and recall, the number of false positives that we get is rather high in relation to our universe. As we have earlier said, a motivating factor of this paper is the automatic induction of the structures that we have called messages; the extracted lexicon of verbs and messages will later be used by an unsupervised clustering algorithm in order to create the classes of words which will correspond to the message types. For this reason, although we prefer to have an F-measure as high as possible, we also want to have a fallout measure as low as possible, so that the number of false positives will not perturb the clustering algorithm.

If, on the other hand, we examine the relation between the F-measure and Fallout, we notice that for the Inter-document Frequency with a threshold value of 4 we obtain a Precision of 71.60\%, a recall of 43.86\% and an F-measure of 54.40\%. Most importantly though we get a fallout measure of 8.86\% which implies that the percentage of wrongly classified verbs and nouns compose a small percentage of the total universe of verbs and nouns. This combination of high F-measure and very low Fallout is very important for later stages during the process of message induction.

\section{Discussion}\label{sec:discussion}
As we have claimed in the introduction of this paper, although we have applied our series of experiments in a single domain, that of terroristic incidents which involve hostages, we believe that the proposed procedure can be viewed as a ``general'' one. In the section we would like to clarify what exactly we mean by this statement.

In order to proceed, we would like to suggest that one can view two different kinds of generalization for the proposed procedure:
\begin{enumerate}
  \item The proposed procedure is a general one in the sense that it can be applied in a large corpus of \emph{heterogeneous} documents incorporating various domains and genres, in order to yield ``general'', \ie domain-independent, frames that can later be used for any kind of domain.
  \item The proposed procedure is a general one in the sense that it can be used in any kind of domain without any modifications. In contrast with the first point, in this case the documents to which the proposed procedure will be applied ought to be \emph{homogeneous} and rather representative of the domain. The induced frames will not be general ones, but instead will be domain dependent ones.
\end{enumerate}

Given the above two definitions of generality, we could say that the procedure proposed in this paper falls rather in the second category than in the first one. Ignoring for the moment the second stage of the procedure---clustering of word sequences characteristic of specific semantic roles---and focusing on the actual work described in this paper, that is the use of statistical methods for the identification of candidate predicates, it becomes clear that the use of an heterogeneous, non-balanced corpus is prone to skewing the results. By consequence, we believe that the proposed procedure is general in the sense that we can use it for any kind of domain which is described by an homogeneous corpus of documents.

\section{Related Work}\label{sec:related_work}
Teufel and Moens \shortcite{teufelAl02} and Saggion and Lapalme \shortcite{saggionAl02} have shown that templates based on domain concepts and relations descriptions can be used for the task of automatic text summarization. The drawback of their work is that they rely on  manual acquisition of lexical resources and semantic classes' definition. Consequently, they do not avoid the time-consuming task of elaborating linguistic resources. It is actually for this kind of reason---that is, the laborious manual work---that automatic induction of various structures is a recurrent theme in different research areas of Natural Language Processing.

An example of an inductive Information Extraction algorithm is the one presented by Fabio Ciravegna \shortcite{Ciravegna.01}. The algorithm is called \textsc{(lp)$^2$}. The goal of the algorithm is to induce several symbolic rules given as input previous SGML tagged information by the user. The induced rules will later be applied in new texts in order to tag it with the appropriate SGML tags. The induced rules by \textsc{(lp)$^2$} fall into two distinct categories. In the first we have a bottom up procedure which generalizes the tag instances found in the training corpus which uses shallow NLP knowledge. A second set of rules is also created which have a corrective character; that is, the application of this second set of rules aims at correcting several of the mistakes that are performed by the first set of rules.

On the other hand several researchers have pioneered the automatic acquisition of lexical and semantic resources (such as verb classes). Some approaches are based on Harris's \shortcite{harris51} distribution hypothesis: syntactic structures with high occurrences can be used for identifying word clusters with common contexts \cite{pantelAl01}. Some others perform analysis from semantic networks \cite{resnikAl04}. Poibeau and Dutoit \shortcite{poibeauAl02} showed that both can be used in a complementary way.

Currently, our approach follows the first trend. Based on Hernandez and Grau~\shortcite{hernandezAl03,hernandez04b}'s proposal, we aim at explicitly using corpus characteristics such as its genre and domain features to reduce the quantity of considered data. 
%
%
In this paper we have explored various statistical measures which could be used as a filter for improving results obtained by the previous mentioned works.


\section{Conclusions and Future Work}\label{sec:conclusions}
In this paper we have presented a statistical approach for the extraction of a lexicon which contains the verbs and nouns that can be considered as candidates for use as predicates for the induction of predicate/argument structures that we call messages. Actually, the research presented here can be considered as the first step in a two-stages approach. The next step involves the use of clustering algorithms on the extracted lexicon which will provide the final clusters that will contain the predicates and arguments for the messages. This process is itself part of a larger process for the induction of predicate/argument structures. Apart from messages, such structures could as well be the structures that are associated with frame semantics, that is the frames and their associated semantic roles. Despite the great resemblances that messages and frames have, one of their great differences is the fact that messages were firstly introduced in the context of automatic multi-document summarization. By consequence they are meant to capture the most important information in a domain. Frames and semantic roles on the other hand, do not have this restriction and thus are more general. Nonetheless, it is our hope that the current research could ultimately be useful for the induction of frame semantics. In fact it is in our plans for the immediate future work to apply the same procedure in \fn annotated data\footnote{See \url{http://framenet.icsi.berkeley.edu/index.php?option=com_wrapper&Itemid=84}} in order to extract a vocabulary of verbs and nouns which will be characteristic of the different Linguistic Units (LUs) for the frames of \fn.

The proposed statistical measures are meant to be a first step towards a fully automated process of message induction. The immediate next step in the process involves the application of various unsupervised clustering techniques on the obtained lexicon in order to create the 48 different classes each one of which will represent a distinct vocabulary for the 48 different message types. We are currently experimenting with several algorithms such \emph{K-means, Expectation-Minimization (EM), Cobweb} and \emph{Farthest First}. In addition to those clustering algorithms, we are also examining the use of various lexical association measures such as \emph{Mutual Information, Dice coefficient, $\chi^2$}, etc. Although this approach will provide us with clusters of predicates and candidate arguments, still the problem of linking the predicates with their arguments remains. Undoubtedly, the use of more linguistically oriented techniques, such as syntactic analysis, is inevitable. We are currently experimenting with the use of a shallow parser (chunker) in order to identify the chunks that behave similarly in respect to a given cluster of predicates.

Concerning the evaluation of our approach, the highest F-measure score (61,87\%) was given by the Collection Frequency statistical measure with a threshold value of 46\%. This high F-measure though came at the cost of a high Fallout score (31.16\%). Since the extracted lexicon will later be used as an input to a clustering algorithm, we would like to minimize as much as possible the false positives. By consequence we have opted in using the Inter-document Frequency measure which presents an F-measure of 54.40\% and a much more limited Fallout of 8.86\%.

\small
\bigskip\noindent\textbf{Acknowledgments} \ \\
The authors would like to thank Konstantina Liontou and Maria Salapata for their help on the annotation of the messages, as well as the anonymous reviewers for their insightful and constructive comments.
\normalsize

\bibliographystyle{acl}
\bibliography{CACLA_EACL09}
\end{document}